\documentclass[11pt,a4paper]{article}
\usepackage[hyperref]{acl2021}
\usepackage{times}
\usepackage{latexsym}

\usepackage{microtype}

\aclfinalcopy % Uncomment this line for the final submission
\usepackage{beamerarticle}
\usepackage{latexsym}
\usepackage{microtype}
\usepackage{booktabs}
\usepackage{url}
\usepackage{eurosym}
\usepackage{todonotes}
\usepackage{multirow}
\usepackage{subcaption}
\usepackage{amssymb}
\usepackage{amsmath}
\usepackage{mathtools}
\usepackage{array}
\usepackage{pgfplotstable}
\usepackage{pgfplots}
\usepackage{soul}
\usepackage{xcolor}
\usepackage{txfonts}
\usepackage{pgf}
\usepackage{tikz}
\usepackage{adjustbox}
\usepackage{appendix}
\usetikzlibrary{decorations.pathmorphing,patterns}
\usepgfplotslibrary{groupplots,units}
\usetikzlibrary{backgrounds,positioning,fit,shapes}
\pgfplotsset{compat=newest}

\newif\ifcomments
\commentstrue
\ifcomments
    \providecommand\matt[1]{\textcolor{teal}{[Matt: #1]}}
    \providecommand\nitish[1]{\textcolor{violet}{[Nitish: {#1}]}}
    \providecommand\dheeru[1]{\textcolor{violet}{[Dheeru: {#1}]}}
    \providecommand\pradeep[1]{\textcolor{violet}{[Pradeep: {#1}]}}
    \providecommand\sameer[1]{\textcolor{purple}{[Sameer: #1]}}
\else
    \providecommand{\matt}[1]{}
    \providecommand{\nitish}[1]{}
    \providecommand{\dheeru}[1]{}
    \providecommand{\pradeep}[1]{}
    \providecommand{\sameer}[1]{}
\fi

\title{Learning with Instance Bundles for Reading Comprehension}

\author{Dheeru Dua\textsuperscript{$\clubsuit$}, Pradeep Dasigi\textsuperscript{$\spadesuit$}, \\ \textbf{Sameer Singh}\textsuperscript{$\clubsuit$}, and \textbf{Matt Gardner}\textsuperscript{$\spadesuit$} \\
  \textsuperscript{$\clubsuit$}University of California, Irvine, USA \\
  \textsuperscript{$\spadesuit$}Allen Institute for Artificial Intelligence \\
  {\tt ddua@uci.edu} \\}

\date{}

\begin{document}
\maketitle
\begin{abstract}
When training most modern reading comprehension models, all the questions associated with a context are treated as being independent from each other. However, closely related questions and their corresponding answers are not independent, and leveraging these relationships could provide a strong supervision signal to a model.  Drawing on ideas from contrastive estimation, we introduce several new supervision techniques that compare question-answer scores across multiple related instances.  Specifically, we normalize these scores across various neighborhoods of closely contrasting questions and/or answers, adding another cross entropy loss term that is used in addition to traditional maximum likelihood estimation.  Our techniques require bundles of related question-answer pairs, which we can either mine from within existing data or create using various automated heuristics.   We empirically demonstrate the effectiveness of training with instance bundles on two datasets---HotpotQA and ROPES---showing up to 11\% absolute gains in accuracy.

%In most modern reading comprehension datasets, all the questions associated with a context are treated as being independent from each other. However, complex natural language tasks like reading comprehension can benefit from learning how to answer a set of closely related questions concurrently.
%Existing methods learn interactions between a single question and multiple candidate answers, but none model interactions across question-answer pairs.
%We introduce a novel energy-based technique that compares question-answer scores across multiple related questions, not just across multiple answers for the same question, which can be seen as regularizing the energy-based model to permit several different probabilistic interpretations.
%Our technique requires bundles of related question-answer pairs, which we create using various automated heuristics. 
%We give a thorough investigation of existing techniques, showing for the first time that these methods give substantial performance improvements in reading comprehension settings.  Our novel technique that uses instance bundles shows further gains on top of these existing methods, giving up to 11\% performance increases on two datasets: HotpotQA and ROPES. Finally, we also show how these bundles, if available at test time, can be used to improve the overall answering performance. 
\end{abstract}

%\begin{abstract}
    %Recent works on contrast sets have shown that reading comprehension models can be inconsistent in answering minimally different (related) questions. We conjecture that this is because the training instances, even though related, are assumed to be drawn from an independent and identical distribution. In this work, we study various methods to learn from a bundle of related QA pairs concurrently. The paper describes three major contributions. First, we experiment with a range of automated heuristics to create bundles of related QA pairs. Second, we introduce different compatibility functions that score a given QA pair. We further use these scores to develop a gamut of energy-based probabilistic models. Finally, we show empirically that training concurrently in this manner results in up to 11\% improvement in exact match accuracy on two datasets: HotpotQA and ROPES.
%\end{abstract}

% \input{ACL2021/1-introduction}
\section{Introduction}
Machine learning models are typically trained with the assumption that the training instances sampled from some data distribution are independent and identically distributed.  However, this assumption can cause the learner to ignore distinguishing cues~\cite{dietterich1997solving} between related or minimally different
questions associated with a given context, resulting in inconsistent model learning~\cite{asai-hajishirzi-2020-logic,jia2017adversarial}. 
In a dataset like ROPES, where the ideology of collecting pairs of minimally different questions is taken to its extreme, we see that the performance of a competitive baseline model (RoBERTA) is close to random~\cite{lin2019reasoning}. 
One potential reason for this poor performance is that the model considers each question independently, instead of looking at differences between related questions.

\begin{figure}[!t]
\begin{subfigure}[b]{0.5\textwidth}
\begin{adjustbox}{width=\linewidth}
\begin{tikzpicture}
\small
\node[draw,text width=10cm]  at (0, 0) {\normalsize \textbf{Context:} Marsilea is a genus of approximately 65 species of aquatic ferns of the family Marsileaceae. The name honours Italian naturalist Luigi Ferdinando Marsili (1656-1730) .... Brabejum is a genus of a single species of large evergreen tree, Brabejum stellatifolium in the Proteaceae, commonly called wild almond, bitter almond or ghoeboontjie. \\
\vspace{0.2cm}
\textbf{Question 1:} Is the Marsilea or the Brabejum the genus of more individual species of plants? \\
\textbf{Answer 1:} Marsilea \\
\vspace{0.2cm}
\textbf{Question 2:} Is the Marsilea or the Brabejum the genus of less individual species of plants? \\
\textbf{Answer 2:} Brabejum \\
};
\end{tikzpicture}
\end{adjustbox}
\caption{Instance bundle created from HotpotQA}
\label{fig:hotpot_ex_bundle}
\end{subfigure}

\vspace{0.2cm}

\begin{subfigure}[b]{0.5\textwidth}
\includegraphics[width=\linewidth,height=6.5cm]{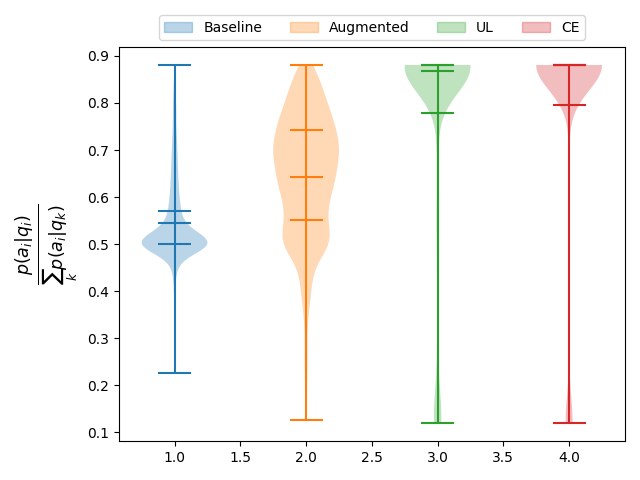}
% \begin{adjustbox}{width=\linewidth}
% \begin{tikzpicture}
% \small
% \pgfplotstableread{
%         pr Baseline Aug. UL CE
%         Marsilea($A_1$) -15.09010692 -1.50124673 2.97124215 3.85630326
%         Brabejum($A_2$) -3.20814397 1.03800033 3.05807351 3.87529105
% }\data
% \begin{groupplot}[
% group style={group size= 1 by 1},
% height=8cm,width=7.4cm,ybar=2pt,
% ymin=-9,ymax=5,xtick=data, enlarge x limits=0.4,
% tick label style={font=\normalsize},
% symbolic x coords={Marsilea($A_1$),Brabejum($A_2$)}, 
% ylabel style={align=center},
% restrict y to domain*=-8:5,
% visualization depends on=rawy\as\rawy,
% after end axis/.code={ % Draw line indicating break
%     \draw [ultra thick, white, decoration={snake, amplitude=1pt}, decorate] (rel axis cs:0.07,0.2) -- (rel axis cs:0.18,0.2);
% },
% nodes near coords={
%     \scriptsize\pgfmathprintnumber{\rawy}
% },
% clip=false,
% legend style={at={(0.7,0.15), font=\tiny},
% anchor=north,legend columns=-1},
% legend image code/.code={
%         \draw [#1] (0cm,-0.1cm) rectangle (0.2cm,0.2cm); }
% ]

% \nextgroupplot[title=,ylabel=$\log\log \frac{p(A_i|Q_i)}{p(A_i|Q_{\setminus i})}$, bar width=10pt]
%     \addplot[fill=red]  table[x=pr,y=Baseline,col sep=space]  {\data};
%     \addplot[fill=gray]  table[x=pr,y=Aug.,col sep=space]  {\data};
%     \addplot[fill=teal]  table[x=pr,y=UL,col sep=space]  {\data};
%     \addplot[fill=brown]  table[x=pr,y=CE,col sep=space]  {\data};
% \legend{Baseline,Aug.,UL,CE}
% \end{groupplot}
% \end{tikzpicture}
% \end{adjustbox}
\caption{Probability of gold QA pair normalized over all questions in the bundle. The higher value indicates that positive QA pair has a high likelihood and at the same time negative QA pair has a low likelihood. At 0.5, both the contrastive questions would produce the same answer with the same likelihood. In an ideal scenario, the distribution should be a delta function at 1.0}
\label{fig:hotpot_ex_prob}
\end{subfigure}
\label{fig:hotpot_ex}
\end{figure}

To address this problem, we propose to train models with sets of related question-answer (QA) pairs simultaneously, instead of having a loss function that decomposes over independent examples.
We use the term \emph{instance bundle} to refer to these sets of closely contrasting examples. Consider an instance bundle from HotpotQA in Figure~\ref{fig:hotpot_ex_bundle}, containing two contrastive QA pairs, which differ in their input by only one word (changing \emph{more} to \emph{less}), resulting in different answers. With both of these examples in a training set, a model trained with traditional maximum likelihood estimation will be incentivized to figure out the difference between their inputs that leads to the expected difference between their answers, but the instances are likely to be seen far apart from each other during training, giving only a weak and indirect signal about their relationship. 

% \dheeru{Even if they are seen in same minibatch (close), I think the average update would still try to increase likelihood of a0 (for q0) at the cost of all possible answers and not just a1}

In order to more effectively learn from these instance bundles, we draw on contrastive estimation~\cite{smith-eisner-2005-contrastive}, a method for re-normalizing an unsupervised probabilistic model using a neighborhood of related examples (originally a set of perturbations of some observed text).  We extend this technique to apply to supervised reading comprehension problems by carefully selecting appropriate ``neighborhoods'' from instance bundles.  The simplest choice of neighborhood is the set of contrasting answers from the instance bundle, resulting in a method similar to unlikelihood training~\cite{welleck2019neural}  or noise-contrastive estimation~\cite{gutmann2010noise}.  However, there are other choices, including the set of contrasting \emph{questions}, or combinations of questions and answers.  These re-normalized loss functions are not effective on their own, which is likely why they have not been used before for training reading comprehension models, but when combined with maximum likelihood training they give substantial increases in performance.  

An intuitive explanation of the reason for this performance improvement is shown in Figure~\ref{fig:hotpot_ex_prob}.  When trained on non-contrasting data with maximum likelihood estimation, a model gives roughly equal values for both $p(A_1|Q_1)$ and $p(A_1|Q_2)$, even though $Q_1$ and $Q_2$ are in some sense opposites.  Adding the contrasting data helps the model differentiate these two probabilities, but not as much as unlikelihood training, which itself is not as effective as contrastive estimation.

We empirically demonstrate the utility of this approach on two reading comprehension datasets: HotpotQA~\cite{yang2018hotpotqa} and ROPES~\cite{lin2019reasoning}.  In order to use instance bundles on these datasets, we introduce various heuristics for obtaining closely related instances.  We show that using contrastive estimation on the instance bundles that we obtain gives up to an 11\% absolute performance improvement over prior training techniques.  These results strongly suggest that data should be collected in instance bundles wherever possible, to allow for stronger supervision signals during training.
\section{Contrastive Estimation for Reading Comprehension}
\label{sec:setup}
Reading comprehension is the task of producing an answer $a$ given a question $q$ about a context $c$.  The question is tied to a particular passage, so in the discussion that follows we will typically use $q$ as a shorthand to refer to both $q$ and $c$ together. Reading comprehension models are typically trained to maximize the likelihood of the answer to each training question.  Given a model's exponentiated scoring function $\psi(q, a)$ for a QA pair,\footnote{$\psi$ is parameterized by model parameters $\theta$, but we omit this in the equations for simplicity of exposition.} this objective normalizes the scores over all possible answer candidates $\mathcal{A}$ for a given question:
\begin{equation*}
    \begin{split}
       \mathcal{L}_\text{MLE}(q_i, a_i) &= \log p(a_i|q_i) \\
       & = \log \frac{\psi(q_i, a_i)}{\sum_{c \in \mathcal{A}} \psi(q_i, c)}
    \end{split}
\end{equation*}

In this work we use a generative model for $\psi$, but many other alternatives are available, and our contribution is applicable to any scoring function. Specifically, we use as $\psi$ the (locally normalized) probability assigned by the generative model to an answer candidate for a given question.

Instead of normalizing scores over all possible answer candidates, \emph{contrastive estimation}~\cite[CE;][]{smith-eisner-2005-contrastive} normalizes scores over some \emph{neighborhood} of closely related instances.  This method was originally introduced for unsupervised linguistic structure prediction, with a neighborhood obtained by permuting observed text to get inputs that had similar content but were ungrammatical.  Our contribution is to apply this general idea to supervised reading comprehension problems.  In our setting, given a neighborhood $\mathcal{N}(q, a)$ of related QA pairs, CE can be described as
\begin{equation*}
        \mathcal{L}_\text{CE}(q_i, a_i) =
        \log \frac{\psi(q_i, a_i)}{\sum\limits_{q_j, a_k \in \mathcal{N}(q_i, a_i)} \psi(q_j, a_k)}
    \label{eq:ce_gen}
\end{equation*}

\citet{smith-eisner-2005-contrastive} \emph{replaced} the MLE objective with CE, which worked well in their unsupervised learning problem. In supervised learning, MLE is a much stronger training signal, and CE on its own severely underperforms MLE. This is because CE provides no learning signal for the very large space of alternative answers to a question that are not in the neighborhood.  However, CE can provide a much stronger signal than MLE for a small set of potentially confusing alternatives, as there are fewer ways for the model to increase the probability of the correct answer.  To adapt CE to supervised settings we interpolate between the two losses, instead of replacing MLE with CE:

\begin{equation*}
    \mathcal{L} = \alpha_1 \mathcal{L}_{MLE} + \alpha_2 \mathcal{L}_{CE}
\end{equation*}

Interestingly, this can be viewed as forcing the scoring function $\psi$ to permit multiple different probabilistic interpretations, as both losses perform softmaxes over different sets of alternatives.  Additionally, if $\psi$ has some locally-normalized component, as is true for the generative models we work with and for many other common models (such as BIO tagging, or independent span start and span end positions), this interpolation in some sense trades off between the locally-focused MLE and the more global view of the problem that the normalization in CE provides (see \S\ref{sec:gen_score} for further discussion of this point).

The key question in applying CE to reading comprehension is how to choose a neighborhood $\mathcal{N}$ for a given training example.  We do so by making bundles of related instances, then extracting various combinations of questions and answers from a bundle to use as neighborhood. 
Formally, an \emph{instance bundle} $\mathcal{B}$ is a collection of unique questions $\mathcal{B}_\mathcal{Q}$ and unique answers $\mathcal{B}_\mathcal{A}$, such that there is at least one QA pair where $a$ is the correct answer to $q$: $\mathrm{ans}(q) = a$.  We refer to such pairs as $(q_g, a_g)$ in the discussion that follows.
Our assumption is that the questions in $\mathcal{B}_\mathcal{Q}$ and the answers in $\mathcal{B}_\mathcal{A}$ are related to each other in some way---often they differ in only a single word---though we do not characterize this formally.  However, a good bundle creation procedure is crucial for effective model learning.  We discuss several ways for creating bundles in Section~\ref{sec:bundle}, and discuss the limitations of CE when effective bundles cannot be created in \autoref{sec:close-bundles}. The following section discusses choices of neighborhood functions given an instance bundle.

\subsection{Choosing a neighborhood}
\label{sec:neighbor}

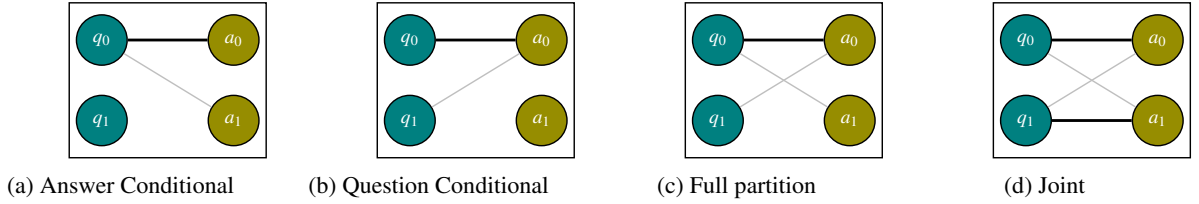
\begin{figure*}
\captionsetup[subfigure]{justification=centering}
\centering
\begin{subfigure}[b]{.24\textwidth}
\begin{adjustbox}{width=\linewidth}
\begin{tikzpicture}[-,>={stealth[black]},node distance=1.6cm,thick]
  \tikzstyle{question}=[circle,draw,fill=teal,text=white,minimum size=1cm]
  \tikzstyle{answer}=[circle,draw,fill=olive,text=white,minimum size=1cm]
  \tikzstyle{latent}=[circle,draw,fill=gray,text=black,minimum size=1cm]
  \tikzstyle{observed}=[circle,draw,fill=gray,text=white,minimum size=1cm]
  \tikzstyle{ellipsis}=[circle,fill=none,text=black,minimum size=1cm]
  \tikzstyle{plate}=[rectangle,draw,fill=none,minimum height=8em, minimum width=5em]
  \tikzstyle{factor} = [rectangle, fill=black,minimum size=5pt, inner
sep=0pt, node distance=0.4]
  \node[ellipsis]           (C)                                 {};
  \node[answer]           (A0) [right of=C, xshift=3cm]       {$a_0$};
  \node[answer]           (A1) [below of=A0, xshift=0cm]      {$a_1$};
  \node[question]         (Q0) [left of=A0, xshift=-1cm]    {$q_0$};
  \node[question]         (Q1) [left of=A1, xshift=-1cm]      {$q_1$};
  \path[black, ultra thick] (Q0) edge              node {} (A0);
  \path[lightgray] (Q0) edge              node {} (A1);
  \node[plate, fit=(Q0)(Q1)(A0)(A1)] (containerq) {};
\end{tikzpicture}
\end{adjustbox}
\caption{Answer Conditional}
\label{fig:ans_cond}
\end{subfigure}
\hfill
\begin{subfigure}[b]{.24\textwidth}
\begin{adjustbox}{width=\linewidth}
\begin{tikzpicture}[-,>={stealth[black]},node distance=1.6cm,thick]
  \tikzstyle{question}=[circle,draw,fill=teal,text=white,minimum size=1cm]
  \tikzstyle{answer}=[circle,draw,fill=olive,text=white,minimum size=1cm]
  \tikzstyle{latent}=[circle,draw,fill=white,text=black,minimum size=1cm]
  \tikzstyle{observed}=[circle,draw,fill=gray,text=white,minimum size=1cm]
  \tikzstyle{ellipsis}=[circle,fill=none,text=black,minimum size=1cm]
  \tikzstyle{plate}=[rectangle,draw,fill=none,minimum height=8em, minimum width=5em]
  \tikzstyle{factor}=[rectangle, fill=black,minimum size=5pt, inner
sep=0pt, node distance=0.4]
  \node[ellipsis]           (C)                                 {};
  \node[answer]           (A0) [right of=C, xshift=3cm]       {$a_0$};
  \node[answer]           (A1) [below of=A0, xshift=0cm]      {$a_1$};
  \node[question]         (Q0) [left of=A0, xshift=-1cm]    {$q_0$};
  \node[question]         (Q1) [left of=A1, xshift=-1cm]      {$q_1$};
  \path[black, ultra thick] (Q0) edge              node {} (A0);
  \path[lightgray] (Q1) edge              node {} (A0);
  \node[plate, fit=(Q0)(Q1)(A0)(A1)] (containerq) {};
\end{tikzpicture}
\end{adjustbox}
\caption{Question Conditional}
\label{fig:ques_cond}
\end{subfigure}
\hfill
\begin{subfigure}[b]{.24\textwidth}
\begin{adjustbox}{width=\linewidth}
\centering
\begin{tikzpicture}[-,>={stealth[black]},node distance=1.6cm,thick]
  \tikzstyle{question}=[circle,draw,fill=teal,text=white,minimum size=1cm]
  \tikzstyle{answer}=[circle,draw,fill=olive,text=white,minimum size=1cm]
  \tikzstyle{latent}=[circle,draw,fill=white,text=black,minimum size=1cm]
  \tikzstyle{ellipsis}=[circle,fill=none,text=black,minimum size=1cm]
  \tikzstyle{observed}=[circle,draw,fill=gray,text=white,minimum size=1cm]
  \tikzstyle{plate}=[rectangle,draw,fill=none,minimum height=8em, minimum width=5em]
  \tikzstyle{factor} = [rectangle, fill=black,minimum size=5pt, inner
sep=0pt, node distance=0.4]
  \node[ellipsis]           (C)                                 {};
  \node[answer]           (A0) [right of=C, xshift=3cm]       {$a_0$};
  \node[answer]           (A1) [below of=A0, xshift=0cm]      {$a_1$};
  \node[question]         (Q0) [left of=A0, xshift=-1cm]    {$q_0$};
  \node[question]         (Q1) [left of=A1, xshift=-1cm]      {$q_1$};
  \path[black, ultra thick] (Q0) edge              node {} (A0);
  \path[lightgray] (Q0) edge              node {} (A1);
  \path[lightgray] (Q1) edge              node {} (A0);
  \node[plate, fit=(Q0)(Q1)(A0)(A1)] (containerq) {};
\end{tikzpicture}
\end{adjustbox}
\caption{Full partition}
\label{fig:full_part}
\end{subfigure}
\hfill
\begin{subfigure}[b]{.24\textwidth}
\begin{adjustbox}{width=\linewidth}
\begin{tikzpicture}[-,>={stealth[black]},node distance=1.6cm,thick]
  \tikzstyle{question}=[circle,draw,fill=teal,text=white,minimum size=1cm]
  \tikzstyle{answer}=[circle,draw,fill=olive,text=white,minimum size=1cm]
  \tikzstyle{latent}=[circle,draw,fill=white,text=black,minimum size=1cm]
  \tikzstyle{observed}=[circle,draw,fill=gray,text=white,minimum size=1cm]
  \tikzstyle{ellipsis}=[circle,fill=none,text=black,minimum size=1cm]
  \tikzstyle{plate}=[rectangle,draw,fill=none,minimum height=8em, minimum width=5em]
  \tikzstyle{factor} = [rectangle, fill=black,minimum size=5pt, inner
sep=0pt, node distance=0.4]

  \node[ellipsis]           (C)                                 {};
  \node[answer]           (A0) [right of=C, xshift=3cm]       {$a_0$};
  \node[answer]           (A1) [below of=A0, xshift=0cm]      {$a_1$};
  \node[question]           (Q0) [left of=A0, xshift=-1cm]      {$q_0$};
  \node[question]           (Q1) [left of=A1, xshift=-1cm]      {$q_1$};
  \path[black, ultra thick] (Q0) edge              node {} (A0);
  \path[lightgray] (Q0) edge              node {} (A1);
  \path[lightgray] (Q1) edge              node {} (A0);
  \path[black, ultra thick] (Q1) edge              node {} (A1);
  \node[plate, fit=(Q0)(Q1)(A0)(A1)] (containerq) {};
\end{tikzpicture}
\end{adjustbox}
\caption{Joint}
\label{fig:joint}
\end{subfigure}

% \hspace{1.5cm}
% \begin{subfigure}[t]{.2\textwidth}
% \centering
% \begin{tikzpicture}[-,>={stealth[black]},node distance=1.6cm,thick]
%   \tikzstyle{question}=[circle,draw,fill=teal,text=white,minimum size=1cm]
%   \tikzstyle{answer}=[circle,draw,fill=olive,text=white,minimum size=1cm]
%   \tikzstyle{latent}=[circle,draw,fill=white,text=black,minimum size=1cm]
%   \tikzstyle{ellipsis}=[circle,fill=none,text=black,minimum size=1cm]
%   \tikzstyle{plate}=[rectangle,draw,fill=none,minimum height=8em, minimum width=4em]

%   \node[ellipsis]           (C)                                 {};
%   \node[answer]           (A0) [right of=C, xshift=3cm]       {$q_0a_1$};
%   \node[answer]           (A1) [below of=A0, xshift=0cm]      {$q_1a_0$};
%   \node[question]           (Q0) [left of=A0, xshift=-1cm]      {$q_0a_0$};
%   \node[question]           (Q1) [left of=A1, xshift=-1cm]      {$q_1a_1$};
%   \path[blue] (Q0) edge              node {} (A0);
%   \path[red] (Q1) edge              node {} (A0);
%   \path[brown] (Q0) edge              node {} (A1);
%   \path[gray] (Q1) edge              node {} (A1);
%   \path[teal] (Q0) edge              node {} (Q1);
%   \path[olive] (A1) edge              node {} (A0);
%   \node[plate, fit=(Q0)(Q1)(A0)(A1)] (containerq) {};
% \end{tikzpicture}
% \caption{Joint(\dheeru{need to figure out how to show this correctly}}
% \label{fig:joint_cond}
% \end{subfigure}
\caption{Contrastive Estimation models. In each subfigure an instance bundle of size 2 is shown, with bold lines indicating combinations whose probability is maximized at the expense of the combinations represented by gray lines, for the positive QA pair $(q_0,a_0)$ in the bundle. The total CE loss is the sum of loss for each positive QA pair in the bundle.}
\label{fig:multi_var}
\end{figure*}

Given an instance bundle $\mathcal{B}$ with questions $\mathcal{B}_\mathcal{Q}$ and answers $\mathcal{B}_\mathcal{A}$, there are many ways to construct a neighborhood.  \autoref{fig:multi_var} shows some of these options graphically, with the bold line showing the gold QA pair, and gray lines showing the other QA pairs that make up the neighborhood.  We distinguish between two kinds of neighborhood methods. A \emph{single neighborhood} CE model is one that perturbs and normalizes over a single variable, either the question (input) or the answer (output). Similarly, \emph{multiple neighborhood} CE models perturb both variables jointly and normalize over the combinatorial space of both variables. 

\subsubsection{Single neighborhood models}
\label{sec:single}
These models construct neighborhood using the set of contrasting answers or contrasting questions from the instance bundle.

% \matt{You need to clarify what's going on here with sum vs. product and log vs. no log in all of these.  I would think that if you have a sum (e.g., in equations 6 and 7), you should have a log inside the sum.  Right?  Something is off in the math throughout this whole section, and this confusion is behind my question in section 3.3.2.  Might be easier to work this out over slack or in a meeting.}

\paragraph{Answer Conditional:}
This probabilistic model maximizes the probability of the correct answer $a_i$ at the expense of the other answers candidates in the instance bundle $\mathcal{B}_\mathcal{A}$ (\autoref{fig:ans_cond}). 
\begin{equation*}
    \begin{split}
        \mathcal{L}_{\text{CE-AC}}(q_g, a_g, \mathcal{B}) = \log \frac{\psi(q_g, a_g)}{\sum_{a_j \in \mathcal{B}_\mathcal{A}} \psi(q_g, a_j)}
    \end{split}
    \label{eq:ans_cond}
\end{equation*}

\paragraph{Question Conditional}
This model computes the normalization constant over the question neighborhood for a fixed answer. This effectively computes a probability distribution over \emph{questions} in the bundle given the correct answer, and maximizes the probability of the correct question (\autoref{fig:ques_cond}).
\begin{equation*}
    \begin{split}
       \mathcal{L}_{\text{CE-QC}}(q_g, a_g, \mathcal{B}) = \log \frac{\psi(q_g, a_g)}{\sum_{q_j \in \mathcal{B}_{\mathcal{Q}}} \psi(q_j, a_g)}
    \end{split}
    \label{eq:ques_cond}
\end{equation*}

\subsubsection{Multiple neighbourhood models}
\label{sec:multi}
These models consider all possible combinations of questions,  $\mathcal{B}_{\mathcal{Q}}$ and answers, $\mathcal{B}_{\mathcal{A}}$ in a bundle for normalization, unlike single neighborhood models which only look at either $\mathcal{B}_{\mathcal{A}}$ or $\mathcal{B}_{\mathcal{Q}}$.

\paragraph{Two Way}
This method simply does a weighted combination~\cite{jacobs1991adaptive} of the answer conditional and question conditional losses.
\begin{equation*}
    \begin{split}
       \mathcal{L}_{\text{CE-TW}}(q_g, a_g, \mathcal{B}) = \lambda_1 \log \frac{\psi(q_g, a_g)}{\sum_{a_j \in \mathcal{B}_\mathcal{A}} \psi(q_g, a_j)} + \\
       \lambda_2 \log \frac{\psi(q_g, a_g)}{\sum_{q_j \in \mathcal{B}_\mathcal{Q}} \psi(q_j, a_g)}
    \end{split}
    \label{eq:two_way_cond}
\end{equation*}

\paragraph{Full Partition}
Instead of separate normalizations over questions and answers, this method does a single normalization over the same sets as in Two Way. This is equivalent to normalizing over the cross product $\mathcal{B}_\mathcal{Q} \times \mathcal{B}_\mathcal{A}$, minus other correct pairings (\autoref{fig:full_part}).

\begin{equation*}
    \begin{split}
       \mathcal{L}_{\text{CE-FP}}(q_g, a_g, \mathcal{B}) = \log \frac{\psi(q_g, a_g)}{\psi(q_g, a_g)+\sum\limits_{\substack{q_j \in \mathcal{B}_{\mathcal{Q}}, \\ a_k \in \mathcal{B}_{\mathcal{A}}, \\ \mathrm{ans}(q_j) \neq a_k}} \psi(q_j, a_k)}
    \end{split}
    \label{eq:full_part}
\end{equation*}

%\paragraph{Multi-label}
%In this model the likelihood of all gold QA pairs is maximized at the same time, similar to multi-label classification. This model can, unfortunately, increase the probability mass of a positive QA pair at the cost of another positive QA pair in the bundle.
%In this model the neighborhood consists of the cross product of $\mathcal{B}_\mathcal{Q}$ and $\mathcal{B}_\mathcal{A}$. This model can, unfortunately, increase the probability mass of a positive QA pair at the cost of another positive QA pair in the bundle.
%\begin{equation}
%    \begin{split}
%       \text{CE-ML}(\mathcal{B}) = \sum_{i=1}^n \log \frac{\psi(q_i, a_i)}{\sum\limits_{\substack{q_j \in \mathcal{Q}_{\mathcal{B}}, \\ a_k \in \mathcal{A}_{\mathcal{B}}}} \psi(q_j, a_k)}
%    \end{split}
%    \label{eq:multilabel}
%\end{equation}

% \matt{If I'm understanding the math right, this formula is wrong.  Shouldn't you have more than one term in the numerator?} \dheeru{ for n = 2 it will become $\frac{\psi(q_i,a_i)\psi(q_j,a_j)}{Z^2}$, its independently increasing lik. of each label, did I misunderstand something?}

\paragraph{Joint}
This method switches from optimizing the probability of single QA pairs to optimizing the \emph{set} of correct QA pairs in the bundle, also known as power-set label classification~\cite{zhang2007ml} (\autoref{fig:joint}). We perform this for only bundles consisting of two correct QA pairs, because the power set becomes prohibitively large for larger bundles.  Let $C(\mathcal{B})$ be a function that returns all unique subsets of size 2 from the cross product set $\mathcal{B}_\mathcal{Q} \times \mathcal{B}_\mathcal{A}$, and let $(q_{g_1}, a_{g_1})$ and $(q_{g_2}, a_{g_2})$ be the two positive QA pairs in the bundle.

The joint CE objective is

\begin{equation*}
    \mathcal{L}_{\text{CE-JT}}(\mathcal{B}) =
    \frac{\psi(q_{g_1}, a_{g_1})\psi(q_{g_2}, a_{g_2})}{\sum\limits_{q_i, a_k, q_j, a_l \in C(\mathcal{B})} \psi(q_i, a_k)\psi(q_j, a_l)}
\label{eq:joint}
\end{equation*}

% \begin{lemma}
% The CE-ML objective is a lower bound on CE-JT objective. 
% \end{lemma}
% \begin{proof}
% The CE-ML objective for independently aligning the right question-answer pair is 
% \begin{equation}
%     \begin{split}
%         \exp(\text{CE-ML}(\mathcal{B})) &= \frac{s_{00}s_{11}}{(s_{00} + s_{11} + s_{10} + s_{01})^2} \\
%         &= \frac{s_{00} s_{11}}{\splitfrac{2(s_{00} s_{11} + s_{00} s_{01} + s_{00} s_{10} \strut}
%         {\splitfrac{{} + s_{01} s_{10} + s_{01} s_{11} + s_{10} s_{11}) } {\splitfrac{+ s_{00}^2 + s_{01}^2 + s_{10}^2 + s_{11}^2}{}}}} \\
%     \end{split}
% \label{eq:lemma}
% \end{equation}

%  $s_{00}^2 + s_{01}^2 + s_{10}^2 + s_{11}^2$ is a positive number. The denominator in Eq.~\ref{eq:lemma} is approximately equal to twice the denominator of Eq.~\ref{eq:joint}. This makes the gradient update for the positive pairs in case of joint model larger than that of multi label classifer.
% \end{proof}

\subsection{Alternative uses of bundles}
\label{sec:alternatives}
Here we briefly consider other potential baselines that make use of instance bundles in some way; we empirically compare against those that are applicable in \autoref{sec:experiments}.

\paragraph{Data Augmentation}
If the bundle $\mathcal{B}$ contains instances that were not present in the training data (e.g., the bundle could be generated using simple heuristics; see \S\ref{sec:bundle}), the simplest use of the bundle is to add all instances to the training data and use MLE under the standard i.i.d. assumption.  This is the standard approach to using this kind of data, and it has been done numerous times previously ~\cite{andreas2019good, zmigrod2019counterfactual}.  This is not applicable if the bundle was obtained by mining the existing training instances, however.

\paragraph{Unlikelihood}
Unlikelihood training~\cite{welleck2019neural} minimizes the likelihood of carefully chosen negative examples to improve a text generation model that would otherwise assign those examples too high of a probability.  Essentially, because the generative model only gets a single positive sequence in an exponentially large set, it does not get strong enough evidence to push down the probability of particularly bad generations.  Unlikelihood training seeks to solve the same problem that contrastive estimation solves, and it provides a natural alternative use of instance bundles.  In our setting, unlikelihood training would decrease the likelihood of negative answers in the bundle:
\begin{equation*}
    \begin{split}
        \mathcal{L}_{\text{UL}}(q_g, a_g, \mathcal{B}) = \mathcal{L}_{\text{MLE}}(q_g, a_g) + \sum_{c \in \mathcal{B}_\mathcal{A} \setminus a_g} \log(1 - p(c|q_g))
    \end{split}
    \label{eq:ull}
\end{equation*}

The second term in Eq.~\ref{eq:ull} helps provide additional signal to further reduce the likelihood of neighbouring negative answers, especially when the MLE loss starts to overfit at training~\cite{lewis2018generative}. 

Unlikelihood training, though easy to perform, has two drawbacks. 
First, it independently minimizes the likelihood of the neighborhood, which means that the probability mass is moved from negative QA pairs but may not necessarily move to the positive pair, unlike CE. Second, because it assumes a conditional probabilistic model of $p(a|q)$, it is not clear how to use alternative questions in the bundle with this objective.

% \paragraph{Noise Contrastive Estimation}
% Contrastive Estimation model can also be theorized as an energy based model which tries to push down energy of the oracle input-output pair over negative pairs. Energy based models often utilize noise contrastive estimation~\cite{gutmann2010noise} (NCE) which provides an efficient way of approximating the normalizer over a large space of candidates by learning a binary classifier, $C$, that distinguishes gold pair from negative pairs. 

% \begin{equation}
%     \begin{split}
%     \text{NCE}(a_i, q_i) = \log p(C=1|a_i,q_i) + \\
%     \sum_{(a_j, q_k) \in {\mathcal{A}_{\mathcal{B}} \times \mathcal{Q}_{\mathcal{B}}}} \log p(C=0|a_j,q_k)
%     \end{split}
%     \label{eq:nce_gen}
% \end{equation}

% As we can see from Eq~\ref{eq:ce_gen} an Eq~\ref{eq:nce_gen}, the answer neighbourhood does not take into account false negative input-output pairs combinations, eg. ($q_j, a_j$) which makes standard NCE (and CE) estimates noisy. However, due to the injective assumption, the CE-FP model described in Section~\ref{sec:multi} can deals with this scenario.

% NCE can easily replace CE in our setup similar to ~\cite{yeh2019qainfomax}. However, since the number of input-output pair combinations are $O(n^2)$, in the number of instances in the bundle, $n$ which is quite small in our case we employed CE instead of NCE. 

%\input{ACL2021/3-alternatives}
\section{Bundling Heuristics}
\label{sec:bundle}

In this section we discuss how we obtain instance bundles for use with contrastive estimation and other related baselines.

A naive way to create a bundle would be to exploit the fact that all the questions associated with a context are likely to be related, and simply make bundles consisting of all QA pairs associated with the context. However, this approach poses two problems. First, there could be many questions associated with any particular context, and smaller, more closely-related bundles are more informative. Second, and relatedly, it is likely that bundles obtained this way will have many questions whose answers can be obtained from the bundle by superficial type matching.  For instance, a wh-question starting with ``where" would most likely align with a location type answer.  If this were bundled with a question starting with ``how many'', with an answer that is a number, the bundle would be largely uninformative.  We instead attempt to create bundles with minimally different questions and answers, in several different ways.

\paragraph{Diverse Top-k sampling}
\label{sec:topk_bundling}
We first discuss a method for getting alternative \emph{answers} to a single question.  This will result in a bundle that can only be used with answer conditional CE, as there are no alternative questions in the bundle.  An easy way to get answer candidates is to employ a pre-trained answering model and sample answers from the posterior distribution. However, since the model has seen all the QA pairs while training, it can easily memorize answers, resulting in a low variance, high confidence distribution. In order to achieve diverse answer samples we need to either over-generate and prune out the gold answer from the samples or induce a diversity promoting sampling. We adopt a hybrid sampling strategy where we use nucleus sampling for the first few timesteps (without replacement) and then top-k for the remaining timesteps. This forces the answer generator to consider different starting positions in the passage and then generate the best answer span (of an appropriate length) from the token produced at the first step.

\paragraph{Question Mining}
\label{sec:qm}
Some datasets, such as ROPES, are constructed with very close question pairs already in the data.  When these exist, we can create instance bundles by finding natural pairings from the training set.  To find these pairings, we cluster the questions with a high lexical bag-of-words overlap based on Jaccard index ($\geq$ 0.8), ensuring that each question in the cluster has a unique answer.  In ROPES, these bundles typically result in bundles of two QA pairs that differ in one or a few words.  In HotpotQA, the other dataset we focus on in this work, there are very few such pairings naturally occurring in the dataset, so we resort to heuristics to create them.

% \paragraph{Domain Specific}
% In certain cases, we employ domain knowledge to create bundles from existing independent training instances. For instance, in HotopotQA, we need to perform multi-hop reasoning (bridge, comparison, intersection, conjunction) to first select a pair of contexts and then answer a question. Given the same pair of gold contexts if the answers are different, then the respective QA pairs can be combined in a bundle.

\paragraph{Question Generation}
\label{sec:qg}
HotpotQA has many questions that are phrased as multiple-choice, with answer options given in the question itself.  These multiple choice questions can most often be rephrased to provide QA pairs that can be bundled with the original question.  For instance, given the question, ``Which animal is faster, turtle or hare?", it is straightforward to obtain a minimally different question with the opposite answer: ``Which animal is slower, turtle or hare?". 
We adopt three main heuristics to generate such questions whenever possible, applicable to any dataset that has questions of this kind.  All of these heuristics require identifying the two plausible answer choices from the question, which can be done with reasonably high precision using simple regular expressions.
\begin{enumerate}
    \item We replace superlatives with their contrasting counterparts, e.g., (taller, smaller), (more, less), etc.
    \item We negate the main verb, e.g., played $\rightarrow$ didn't play, by inflecting the verbs.\footnote{https://spacy.io/universe/project/lemminflect/}
    \item We swap the noun phrases being compared in the question, e.g., ``Are rock A's wavelengths shorter or longer than rock B's?" can be used to generate, ``Are rock B's wavelengths shorter or longer than rock A's?"
\end{enumerate}

\section{Experiments}
\label{sec:experiments}
We use an encoder-decoder style T5-large model for all our experiments.
The baseline models in our experiments are the result of fine-tuning the T5 model on the corresponding tasks using the MLE objective. We compare them against models that are further fine-tuned with a combination of MLE and contrastive estimation objectives as described in Section~\ref{sec:neighbor}. That is, when using various instance bundle techniques, we initialize the model with the weights from the fine-tuned MLE model, then continue training with the new loss function.\footnote{To control for the number of optimization steps, we also tried a baseline where we continued fine-tuning an MLE model using the same setup, but this never improved over the original MLE, so we do not include it in the tables.}  The model takes a concatenated context and question as an input to produce an answer output. We use a learning rate of 2e-5 for ROPES and 5e-5 for COMPARISON with lowercased inputs and outputs. We truncate the concatenated context and question up to a length of 650 for ROPES and 850 for COMPARISON.

% \matt{hyperparameter settings?  this seems light on detail for an experiment section} 

In addition to standard metrics on these datasets, we additionally evaluate using a consistency metric.  This metric evaluates to true only if all the questions in a bundle are answered correctly, and is thus a stricter version of EM.

% We use locally normalized log-likelihood as our compatibility function which computes sum log likelihood of each answer token normalized over the entire vocabulary at each time step. 

\subsection{Main results}
We experiment with three datasets: a subset of HotpotQA containing only comparison type of questions (COMPARISON), full HotpotQA and ROPES.   In general, we find that all variants of CE perform substantially better than MLE alone, with question conditional giving small improvements over other CE variants.  All CE models also outperform all UL and data-augmented MLE models.

\paragraph{COMPARISON}
HotpotQA has several different kinds of questions, with the question category labeled in the original data.  We begin by experimenting with the subset labeled as comparison questions, as they lend themselves most naturally to instance bundles.  For these questions, we adopt the question generation strategy to create instance bundles. Table~\ref{fig:comp_tab} shows a comparison of the baseline MLE model (trained on the comparison subset only) with those further fine-tuned with UL and CE over the instance bundles. Also shown is a comparison with further fine-tuning using MLE on the generated QA pairs (+Aug). 

% \begin{table}[htp]
% \centering
%     \small
%     \begin{tabular}{lccc}
%     \toprule
%       &   {EM} & {F1} & {Consistency} \\
%      \midrule
%       MLE (full HotpotQA) & 57.4 & 65.1 & 36.3\\
%       \midrule
%       \textbf{MLE}  & 70.9 & 77.7 &  51.2 \\ 
%       \textbf{+ Aug}  & 73.4 & 80.6 & 76.7  \\
%       \textbf{+ UL} & 75.1 & 82.4 & 85.8\\
%       \textbf{+ Best CE} & 77.4 & 84.7 & 87.3 \\
%     \bottomrule
%     \end{tabular}
%     \caption{COMPARISON dev set performance}
%     \label{fig:comp_tab}
% \end{table}

\begin{table}[htp]
\centering
    \small
    \begin{tabular}{lccc}
    \toprule
      &   {EM} & {F1} & {Consistency} \\
     \midrule
      MLE (full HotpotQA) & 57.4 & 65.1 & 36.3\\
      \midrule
      \textbf{MLE}  & 70.9 & 77.7 &  51.2 \\ 
      \textbf{+ Aug}  & 73.4 & 80.6 & 76.7  \\
      \textbf{+ UL} & 75.1 & 82.4 & 85.8\\
      \midrule
      + Answer Cond. & 76.0 & 83.7 &  \\
      \textbf{+ Question Cond.} & 77.4 & 84.7 & 87.3 \\
      + Two way & 75.5 & 82.7 &  \\
      + Joint & 75.6 & 83.1 &  \\
      + Full Partition & 77.4 & 84.7 &  \\
    \bottomrule
    \end{tabular}
    \caption{COMPARISON dev set performance}
    \label{fig:comp_tab}
\end{table}

Due to unavailability of the code for best model on the HotpotQA dataset, we use a T5-large model trained on the entire HotpotQA as a proxy for state of the art model. Even though this model has a performance of 81.1 F1 on the whole dev set (close to the current SOTA 83.5 on the leaderboard\footnote{https://hotpotqa.github.io/}), on the comparison subset it performs poorly (65.1). Training an MLE model on just this subset reaches 77.7 F1, which is outperformed by unlikelihood training (82.4 F1).  The best CE performance is from the question conditional model, which gets 84.7 F1.

\paragraph{HotpotQA}
% \pradeep{Feels like this should be a separate subsection titled ``Effect of increasing the number of negative samples'', may be after the current 4.3}
% note form dheeru {is this better?}
We additionally experiment with the entire HotpotQA dataset.  Here we use top-k sampling to create instance bundles, where the top-k answer candidates were sampled from the MLE model we use as a baseline. Table~\ref{tab:hotpot_full_ans_cand} shows the performance of the fine-tuned model as we vary the number of answers in $\mathcal{B}_\mathcal{A}$ with CE-AC loss. The overall performance gets better with CE up to $|\mathcal{B}_\mathcal{A}| = 2$, but reduces after that. On a closer examination of the samples, we find that on average we get two distinct answer candidates and the rest of the candidates are ungrammatical word-piece variations of the two distinct candidates (including the oracle answer). These ungrammatical variations provide a noisy signal that hurts model performance.
% \begin{figure}[t]
%      \centering
%         \begin{tikzpicture}[scale=0.8]
%         \begin{axis}[xlabel= Number of negative samples,
%           ylabel=F1,
%           legend style={at={(0.85,0.02)},anchor=south,font=\tiny}]
%         \addplot table [y=ce_hotpot]{plot_data/num_candidates.dat};
%         \addlegendentry{F1}
%         \end{axis}
%         \end{tikzpicture}
%      \caption{F1 performance on full HotpotQA with increasing number(k) of top-k negative answer candidates}
%      \label{fig:hotpot_full_ans_cand}
% \end{figure}

\begin{table}[htp]
\centering
    \small
    \begin{tabular}{lcc}
    \toprule
       & {F1} \\
     \midrule
      \textbf{MLE} &  81.1  \\
      \midrule 
      +CE-AC ($k=1$)  &  82.5  \\
      +CE-AC ($k=2$) &  83.3  \\
      +CE-AC ($k=3$) &  82.1  \\
      +CE-AC ($k=4$) &  81.8 \\
    \bottomrule
    \end{tabular}
    \caption{F1 performance on full HotpotQA dev set with increasing number (k) of top-k negative answer candidates}
    \label{tab:hotpot_full_ans_cand}
\end{table}

\paragraph{ROPES}
Since ROPES already contains minimally different QA pairs, we use question mining to create instance bundles. We use as the most closely comparable prior work the multi-step model of \citet{liu-etal-2020-multi}, which adds a ROPES-specific architecture on top of RoBERTa-large~\cite{liu2019roberta}.\footnote{UnifiedQA~\cite{khashabi2020unifiedqa} also evaluated on ROPES, but they used much more training data from many other datasets, and a much larger model than we experiment with, so their performance is not particularly comparable.} Our baseline MLE model is a generic T5-large model, with fewer parameters (770M vs. 1.5B) and no special architecture.  \autoref{tab:ropes_main} shows that using CE gives almost a 12\% absolute improvement in EM over an MLE model, and a larger than 12\% improvement in consistency, while UL gives only a few point gain.

% The UnifiedQA~\cite{khashabi2020unifiedqa} model that performs better than \textit{SOTA} (74.3 on test set) is trained on 8 different QA datasets with 11 billion parameters.

\begin{table}[htp]
\centering
    \small
    \begin{tabular}{lccc}
    \toprule
      &   {EM} &  {Consistency} \\
      \midrule
      Multi-step~\cite{liu-etal-2020-multi}  & 71.4 & - \\
     \midrule
      \textbf{T5-large MLE} & 65.7 & 52.1   \\
      \textbf{+ UL}  & 68.3 & 55.6  \\
      \midrule
      + Answer Cond. & 74.5 &  \\
      \textbf{+ Question Cond.} & 76.6 & 64.7 \\
      + Two way & 73.5 &   \\
      + Joint & 72.5 &   \\
      + Full Partition & 75.1 &   \\
    \bottomrule
    \end{tabular}
    \caption{ROPES dev set performance}
    \label{tab:ropes_main}
\end{table}

\subsection{Joint Inference}
In cases where we can generate a bundle given only a question (that is, the answer candidates are clear and our heuristics can generate a contrasting question), we can treat test time inference as a hard assignment problem between questions and answers in the generated bundle. We use the scoring function $\psi(q, a)$ to align each question to an answer in the bundle by optimizing objective below:
\begin{align*}
    \begin{split}
      \max \sum_{\substack{a_j \in \mathcal{B}_A, \\ q_i \in \mathcal{B}_Q}} \psi(q_i, a_j) x_{ij} 
    \end{split}
    \\
       \text{s.t.}\qquad & \sum_{j = 0}^{|\mathcal{B}_A|} x_{ij} = 1, \sum_{i = 0}^{|\mathcal{B}_Q|} x_{ij} = 1 
\end{align*}

\begin{figure}
\begin{adjustbox}{width=0.8\linewidth,height=6cm}
\begin{tikzpicture}
\pgfplotstableread{
metric mle aug ul ce
Independent 79 82.3 84.4 85.8
Joint 85.5 87.5 89.1 90.1
}\dataset

\begin{groupplot}[ybar=2pt,
group style={group size= 1 by 1},
height=8cm,width=7.4cm,xtick=data, 
enlarge x limits=0.4,ymax=92,
tick label style={font=\normalsize},
symbolic x coords={Independent, Joint}, 
ylabel style={align=center},
visualization depends on=rawy\as\rawy,
nodes near coords={
    \scriptsize\pgfmathprintnumber{\rawy}
},
nodes near coords style={
                rotate=90,
                anchor=west
},
legend style={at={(0.45,0.98), font=\tiny},
anchor=north,legend columns=-1},
legend image code/.code={
        \draw [#1] (0cm,-0.1cm) rectangle (0.2cm,0.2cm); }
]

\nextgroupplot[bar width=10pt]
\addplot[draw=black,fill=black!20] table[x=metric,y=mle,col sep=space] {\dataset}; 
\addplot[draw=blue,fill=blue!20] table[x=metric,y=aug,col sep=space] {\dataset}; 
\addplot[draw=brown,fill=brown!40] table[x index=0,y index=3] {\dataset}; 
\addplot[draw=red,fill=red!20] table[x index=0,y index=4] {\dataset};
\draw[black,dashed,line width=1pt] (axis description cs:0.3,0.543) -- (axis description cs:0.703,0.543);
\legend{MLE,Aug,UL,CE}
\end{groupplot}
\end{tikzpicture}
\end{adjustbox}
\caption{Performance (F1) on COMPARISON dev with independent prediction versus joint inference.  Joint inference improves all models. The results are on the subset of COMPARISON for which we have paired instances ($\sim$93\%).}
\label{fig:hard_const}
\end{figure}
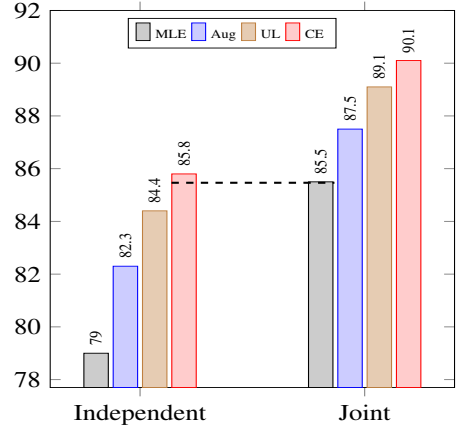

We refer to this as joint inference. Intuitively, even if the model is only given a single question at test time, if it can reason jointly about two competing assignments it can potentially use the alternatives to arrive at a better response than if it only considered the single question it was given.  As shown in \autoref{fig:hard_const}, when using joint inference the performance of a baseline MLE model on COMPARISON improves from 79 F1 to 85.5. The CE model's training paradigm manages to achieve this performance (85.8 F1) without enforcing these constraints at test time, but joint inference additionally improves CE, to 90.1 F1.

\section{Discussion}
In this section we try to understand how CE compares to MLE and UL and under what conditions it is effective.

\subsection{Relation between MLE, UL and CE}
In this section we try to understand the relation between CE, MLE and UL in the special case when the scoring function $\psi$ comes from a locally-normalized generative model, as it does in this work.
Let $p_V$ be the locally normalized probability of an answer candidate (i.e., the combined likelihood of each token in a sequence under a given generative model). $\psi(q, a)$ then equals $p_V(a|q)$.  The CE-AC loss with locally normalized compatibility score can be described as

\begin{align}
    \begin{split}
    \mathcal{L}_{\text{CE-AC}}(q_g, a_g, \mathcal{B}) &=
    \log \frac{ p_V(a_g|q_g)  } {\sum\limits_{c \in \mathcal{B}_\mathcal{A}} p_V(a_c|q_g)} 
    \end{split}
    \label{eq:ce_ac_ln}
\end{align}

We can decompose and rewrite Eq.~\ref{eq:ce_ac_ln} as
\begin{equation}
    \begin{split}
       \mathcal{L}_{\text{CE-AC}}(q_g, a_g, \mathcal{B})  
      &= \log p_V(a_g|q_g) - {\log \sum\limits_{c \in \mathcal{B}_\mathcal{A}} p_V(a_c|q_g)} \\
      &= \mathcal{L}_{MLE}(q_g, a_g) + Reg(\mathcal{B}_\mathcal{A}, q_g)
    \end{split}
    \label{eq:reg}
\end{equation}

Eq.~\ref{eq:reg} shows that $\mathcal{L}_{\text{CE-AC}}$ is just a linear combination of MLE and a regularization term which decreases the probability of each incorrect answer in the bundle.

% Matt: I don't think the following sentence adds much, and "which results in" probably should be "which happens when".  But that probability will never be 1, so it's not clear what the point of the sentence is.
% Dheeru: yeah thats true
% In other words, the $\mathcal{L}_{\text{CE-AC}}$ loss becomes equal to MLE when the regularizer term becomes zero, which results in $\sum_{c \in \mathcal{B}_\mathcal{A}} p_{V}(c|q_i) = 1$.

On a closer look we can see an interesting connection between the regularization term and unlikelihood. The regularization term in CE-AC is essentially the log of an unlikelihood term, except the unlikelihood objective in Section~\ref{sec:alternatives} in practice gets applied at each timestep of decoding, while the regularization term in CE-AC is applied over the entire answer sequence.

Our formulation of CE is more general than the specific case we are analyzing here, but we make note of it as this is the function that we used in our experiments, and it significantly outperformed unlikelihood training.  The theoretical connections shown here could benefit from further exploration.

\subsection{The importance of close instance bundles}
\label{sec:close-bundles}
Experiments on ROPES and COMPARISON show strong improvements by using CE and UL when instances can be grouped into very closely related bundles.  But such effective grouping may not be possible on all datasets. To analyze the applicability of our methods to a dataset without natural bundles, we looked at \textsc{Quoref}~\citep{dasigi-etal-2019-quoref}. Table~\ref{tab:quoref_comparison} shows a comparison between the trends of improvements due to UL and CE across \textsc{Quoref}, ROPES and COMPARISON with bundles created using top-k sampling. As it can be seen from the results, UL does not improve on top of MLE, and CE shows only a very small improvement which is likely statistical noise. To understand why, we analyzed the $p(a | q, c)$ distribution of the baseline MLE model, and computed the following two measures on a random sample of the training set.
\begin{itemize}
    \item $\text{Entropy}_{10} = -\sum_{i=1}^{10} p(a_i | q, c) \log p(a_i | q, c)$
    \item $\text{Top-2 ratio} = \log p(a_1|q,c) / p(a_2|q,c)$ 
\end{itemize}
As seen in Table~\ref{tab:quoref_comparison}, \textsc{Quoref} has a lower $\text{Entropy}_{10}$, and a higher Top-2 ratio than the other datasets, indicating that the baseline MLE model places a lot more weight on the top-1 answer in this task. Manual analysis additionally found that most of the top predictions were ungrammatical variations of the top-1 answer, similar to (but more extreme than) what was seen on the full HotpotQA dataset.  This could explain why the top-k bundling heuristic is not as effective in the case of \textsc{Quoref} as the other two datasets. More generally, these results indicate the importance of effective instance bundling heuristics, and future work could focus on identifying more general ways to create bundles.

\begin{table}
\centering
    \small
    \begin{tabular}{lccccc}
    \toprule
     Dataset & MLE & UL & CE & $\text{Entropy}_{10}$ & Top-2 ratio \\
     \midrule
     COMP. & 77.7 & 82.4 & 84.7 & 2.31 & 0.5 \\
     ROPES & 65.7 & 68.3 & 77.6 & 0.97 & 2.3 \\
     \textsc{Quoref} & 84.8 & 83.9 & 85.0 & 0.06 & 3.1\\
    \bottomrule
    \end{tabular}
    \caption{Comparison between \textsc{Quoref}, COMPARISON, and ROPES datasets with Top-k bundling. MLE, UL, and CE results are on the corresponding development sets ($F_1$ for COMPARISON and \textsc{Quoref}, EM for ROPES) and $\text{Entropy}_{10}$ and Top-2 ratio are measured on random samples of the training sets. UL and CE columns show results after fine-tuning the baseline MLE model with the respective objectives.}
    \label{tab:quoref_comparison}
\end{table}
\section{Related Work}
Learning with negative samples has been explored in many natural language tasks, such as dialogue generation~\cite{cai2020group}, word embeddings~\cite{mikolov2013distributed}, language modeling~\cite{noji2020analysis}, etc., and computer vision tasks such as image captioning~\cite{dai2017contrastive}, unsupervised representation learning~\cite{hadsell2006dimensionality}, etc. In similar vein, mutual information minimization based learners in question answering~\cite{yeh2019qainfomax} and image classification ~\cite{hjelm2018learning} try to decrease the mutual information between positive and negative samples. 

Natural language applications often sample negative examples either randomly from the data or based on likelihood (or unlikelihood) metrics from a reference model. However, the negative samples extracted in this manner are often unrelated.  A growing body of literature is exploring ways to obtain closely-related examples, either manually~\citep{kaushik2020explaining,gardner-etal-2020-evaluating} or automatically~\citep{ribeiro-etal-2020-beyond,Ross2020ExplainingNM,Wu2021PolyjuiceAG}.  This trend is complementary to our work, as we show how to make better use of these closely-related examples during training.  There is also work on consistent cluster assignments in coreference resolution~\cite{chang2011inference}; factually consistent summaries~\cite{kryscinski2019evaluating} and language models~\cite{elazar2021measuring}.

There is also a growing body of literature on training with closely related examples, to which we are contributing.  Several works make use of logical consistency in natural language inference tasks~\citep{Minervini2018AdversariallyRN,li-etal-2019-logic,asai2020logic}.
Another line of work~\cite{teney2019incorporating,teney2020learning,jacovi2021contrastive,Gupta2021PairedEA} tries to increase (or decrease) the distance between intermediate representations of contrasting (or paraphrased) instances.

\section{Conclusion}
We have presented a way to use contrastive estimation in a supervised manner to learn from distinguishing cues between multiple related QA pairs (i.e. instance bundles). Our experiments with multiple CE-based loss functions, defined over a joint neighborhood of questions and answers, have shown that these models outperform existing methods on two datasets: ROPES and HotpotQA. Apart from presenting several ways to create instance bundles, we also explore theoretical connections between unlikelihood training and contrastive estimation, and initial exploration into when instance bundles are likely to be effective with these methods.  We believe our results give strong motivation for further work in techniques to both create and use instance bundles in NLP datasets.

\bibliographystyle{acl_natbib}
\bibliography{anthology,acl2021}

\clearpage

\appendix

\section{Compatabiliy functions for conditional generation models}
\label{sec:gen_score}
We experimented with a few choices of compatibility functions and presented the overall best one in the paper. We describe all the compatibility functions we tried in detail first, and then show results of our experiments on all of them. Our choice of compatibility functions are specific to encoder-decoder style architecture. 

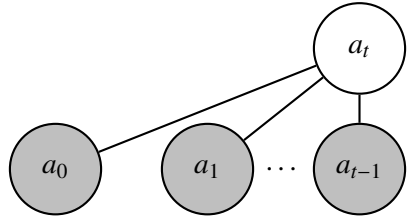
\begin{figure}[t]
\begin{tikzpicture}[-,>={stealth[black]},node distance=1.6cm,thick]
  \tikzstyle{observed}=[circle,draw,fill=lightgray,text=black,minimum size=1.2cm]
  \tikzstyle{latent}=[circle,draw,fill=white,text=black,minimum size=1.2cm]
  \tikzstyle{ellipsis}=[circle,fill=none,text=black,minimum size=1.2cm]

  \node[latent]           (A)                  {$a_t$};
  \node[observed]         (B) [below of=A]     {$a_{t-1}$};
  \node[ellipsis]         (C) [left of=B, xshift=0.6cm]      {\ldots};
  \node[observed]         (D) [left of=C, xshift=0.6cm]      {$a_1$};
  \node[observed]         (E) [left of=D, xshift=-0.4cm]      {$a_0$};
  
  \path (A) edge              node {} (B)
            edge              node {} (D)
            edge              node {} (E);
       
\end{tikzpicture}
\caption{Answer decoder}
\label{fig:answer_decode_graph}
\end{figure}

A transformer style decoder, $d$, can be described as a markov random field~\cite{deng2020cascaded} which takes as input previous answer tokens and question to output current answer token at time step, $t$, as shown in Figure~\ref{fig:answer_decode_graph}. This decoder allows for independent parallel prediction of tokens at each time step, which make the answer likelihood a product of independent markov random fields,  $\prod_t p(a_t|a_{<t},q)$, tied by same parameters in function $d$.

\paragraph{Locally-normalized (LN)} scores can be defined as sum of log-probability of token at each time step locally normalized over vocabulary.
\begin{equation}
    \begin{split}
         \psi_L(q,a) &= \exp \log \prod_t p(a_t|a_{<t}, q)  \\
         &= \exp \sum_{t} \log \frac{\exp(d(a_t; a_{<t},q))}{\sum_{v \in |V|} \exp(d(v; a_{<t},q))}
    \end{split}
\end{equation}

\paragraph{Un-normalized scores (UN):}
Locally normalized scores may suffer from label bias~\cite{goyal2019empirical} which can be crucial when contrastive answers that have overlapping subsequences, for eg., if answer candidates are a list of choices \{``Person A", ``Person B", ``Person C"\}. 

\begin{equation}
    \begin{split}
         \psi_U(q,a) &= 
         \exp \log \prod_{t} \exp(d(a_t; a_{<t},q)) \\
         &= \exp \log \exp(\sum_{t} d(a_t; a_{<t},q)) \\
         &  = \exp \sum_{t} d(a_t; a_{<t},q)
    \end{split}
\end{equation}

\paragraph{ Whole Sequence (GS):} score considers the score of the last token, which is often a special symbol like, $\langle eos \rangle$ tag. Intuitively, this can be seen as a score for the entire input answer sequence, $\psi_G(q, a) = d(a_T; a_{<T},q)$

\subsection{Ablation results}
From Table~\ref{tab:comp_results} and Table~\ref{tab:ropes_results} we can see that the question conditional loss overall works well. Interestingly GS compatibility function performs well on COMPARISON but not on ROPES. We conjecture that's because in ROPES often the candidates have very small lexical dissimilarities (e.g., ``Patient A", ``Patient B") which makes looking at each token important. In ROPES, full partition loss performs slightly better than question conditions. 

\begin{table*}[!htp]
\centering
    \small
    \begin{tabular}{lccccc}
    \toprule
       & \multicolumn{2}{c}{\textbf{Single Neighborhood}} & \multicolumn{3}{c} {\textbf{Multiple Neighborhood}} \\
    \cmidrule(lr){2-3}  \cmidrule(lr){4-6} 
    &   {Answer Conditional} & {Question Conditional} & {Two-way} &   {Joint} & {Full Partition} \\
     \midrule
      \textbf{LN}  & 76.0/83.7 & \textbf{77.4/84.7} & 75.5/82.7  &  75.6/83.1 & 73.2/81.1  \\
      \textbf{UN}  & 75.7/82.9 & 76.6/84.6 & 75.3/82.6 & 75.6/83.3 & 76.7/84.7 \\ 
      \textbf{GS} & 76.9/84.1 & 76.4/83.6 & 76.0/83.2 &  74.7/81.5 & 75.7/83.3\\
    \bottomrule
    \end{tabular}
    \caption{Performance (EM/F1) of COMPARISON dev set on models finetuned with different single and multiple neighborhood CE losses.}
    \label{tab:comp_results}
\end{table*}
\begin{table*}[!htp]
\centering
    \small
    \begin{tabular}{lccccc}
    \toprule
       & \multicolumn{2}{c}{\textbf{Single Neighborhood}} & \multicolumn{3}{c} {\textbf{Multiple Neighborhood}} \\
    \cmidrule(lr){2-3}  \cmidrule(lr){4-6} 
    &   {Answer Conditional} & {Question Conditional} & {Two-way}  & {Joint}  & {Full Partition} \\
     \midrule
      \textbf{LN}  &  74.5 & 76.6 & 73.5  &  72.5 & 75.1\\
      \textbf{UN}  & 73.0 & 75.9 & 74.9  &  71.2 &  \textbf{77.6}\\
      \textbf{GS}  & 71.8 & 69.5 & 71.1 & 69.7 & 73.1\\
    \bottomrule
    \end{tabular}
    \caption{Performance (EM) of ROPES dev set on models finetuned with different single and multiple neighborhood CE losses. We do not report F1 as it's not a good performance measure on this dataset}
    \label{tab:ropes_results}
\end{table*}

% \begin{table*}[!htp]
% \centering
%     \small
%     \begin{tabular}{lcccccc}
%     \toprule
%       & \multicolumn{2}{c}{\textbf{Single Neighborhood}} & \multicolumn{4}{c} {\textbf{Multiple Neighborhood}} \\
%     \cmidrule(lr){2-3}  \cmidrule(lr){4-7} 
%     &   {Answer Conditional} & {Question Conditional} & {Two-way} &  {Multi-label}  & {Joint} & {Full Partition} \\
%      \midrule
%       \textbf{LN}  & 76.0/83.7 & \textbf{77.4/84.7} & 75.5/82.7  & 74.0/80.7 & 75.6/83.1 & 73.2/81.1  \\
%       \textbf{UN}  & 75.7/82.9 & 76.6/84.6 & 75.3/82.6 & 76.0/84.1 & 75.6/83.3 & 76.7/84.7 \\ 
%       \textbf{GS} & 76.9/84.1 & 76.4/83.6 & 76.0/83.2 &  75.6/82.7 & 74.7/81.5 & 75.7/83.3\\
%     \bottomrule
%     \end{tabular}
%     \caption{Comparison EM/F1 contrastive estimation}
%     \label{tab:comp_results}
% \end{table*}
% \begin{table*}[!htp]
% \centering
%     \small
%     \begin{tabular}{lcccccc}
%     \toprule
%       & \multicolumn{2}{c}{\textbf{Single Neighborhood}} & \multicolumn{4}{c} {\textbf{Multiple Neighborhood}} \\
%     \cmidrule(lr){2-3}  \cmidrule(lr){4-7} 
%     &   {Answer Conditional} & {Question Conditional} & {Two-way}  & {Multi-label}  & {Joint}  & {Full Partition} \\
%      \midrule
%       \textbf{LN}  &  74.5 & 76.6 & 73.5  & 69.4 & 72.5 & 75.1\\
%       \textbf{UN}  & 73.0 & 75.9 & 74.9  & 70.2 & 71.2 &  \textbf{77.6}\\
%       \textbf{GS}  & 71.8 & 69.5 & 71.1 & 70.6 & & 73.1\\
%     \bottomrule
%     \end{tabular}
%     \caption{ROPES EM contrastive estimation}
%     \label{tab:ropes_results}
% \end{table*}

\end{document}